\documentclass[12pt]{article} 
\usepackage[english]{babel}
\usepackage[utf8]{inputenc}
\usepackage{johd}

\title{PolInterviews\footnote{The dataset is openly accessible at \url{https://dataverse.harvard.edu/dataset.xhtml?persistentId=doi:10.7910/DVN/03TNGR}} - A Dataset of German Politician Public Broadcast Interviews}

\author{Lukas Birkenmaier$^{a}$$^{*}$, Laureen Sieber $^{b}$, Felix Bergstein$^{c}$ \\
        \small $^{a}$GESIS - Leibniz Institute for the Social Sciences, Mannheim, Germany \\
        \small $^{b}$University of Chemnitz, Chemnitz, Germany \\
        \small $^{c}$University of Mannheim, Mannheim, Germany \\\\
        \small $^{*}$Corresponding author: Lukas Birkenmaier; \tt{lukas.birkenmaier@outlook.de} \\
}
\date{}

\begin{document}
\maketitle
\begin{abstract} 
\noindent This paper presents a novel dataset of public broadcast interviews featuring high-ranking German politicians. The interviews were sourced from YouTube, transcribed, processed for speaker identification, and stored in a tidy and open format. The dataset comprises 99 interviews with 33 different German politicians across five major interview formats, containing a total of 28,146 sentences. As the first of its kind, this dataset offers valuable opportunities for research on various aspects of political communication in the (German) political contexts, such as agenda-setting, interviewer dynamics, or politicians' self-presentation.
\end{abstract}

\noindent\\

\section{Introduction}

Broadcast interviews with politicians, alongside formats like Prime Minister’s questions and political monologues or speeches, serve as central tools for political communication \citep{bull2010}. The primary goal of political communication is persuasion, aiming to shape audience opinions and attitudes through carefully crafted linguistic strategies \citep[2114]{klein2009}. This approach to political representation is largely mediated through the press and digital platforms, a trend \citeauthor{michel2022} (2022: 53 ff.) describes as the “mediatization of politics.” 

Although digital platforms are especially popular among younger audiences, television remains Germany’s most widely accessed medium, achieving the highest daily reach. More than half of Germans rely on television (or television products available on platforms such as YouTube or media libraries) daily to stay informed on national and international political, economic, and cultural events. Online users thus engage with digital content from traditional media sources, including e-papers, websites, apps, and podcasts provided by newspapers and TV stations. The frequency with which these media are used for political communication underscores their substantial influence in shaping public opinion \citep{hein2022}. 

Research primarily relies on debate transcripts to study the content of political interviews. Similarly, modern political science has greatly benefited from advancements in methods for computational text analysis. Tools and algorithms for computational text analysis \citep{grimmer2013text, birkenmaier2023search} have made it possible to systematically analyze large corpora of political texts. However, the applicability of these methods depends heavily on the availability of well-structured, annotated datasets \citep{rauh2020parlspeech, baturo2017understanding}. Collecting such datasets, particularly for specific contexts or languages, remains challenging. In the German context, for instance, there is a notable lack of high-quality transcripts of political interviews altogether. 

Given the central role that professional (digital) content from traditional media plays in the formation of (political) opinion in Germany, this data set is provided. It contains transcripts of 99 journalistic interviews with 33 high-ranking German politicians. The dataset was initially developed to assess whether interview content could reveal strategic personality traits of German politicians (see \citet{birkenmaier2025personality}). However, it is equally suitable for both quantitative and qualitative analyses, including topic-specific or politician-focused studies. Areas of inquiry might include aspects of political communication such as self-presentation, argumentation strategies, or agenda-setting. 

\section{Method}

\subsection{Identification of Interviews} To construct the dataset, we systematically identified interviews with leading German politicians that were publicly available on YouTube. Using domain knowledge, we first identified the most relevant public broadcast formats and inspected their presence on YouTube for the time period between 2020 and 2024. 
We only collected data for high-ranking politicians, which we defined as (co-)partly leaders, secretary generals of parties, and all prime ministers and ministers at both the federal and the country level.
After drafting an initial list of videos, we further applied a backward search for other interviews with the politician. A primary focus was placed on interviews sourced from public broadcasters, where the setting typically involved a one-on-one format with a single politician and one interview host. 
While we made efforts to collect as many relevant interviews as possible, we acknowledge that the dataset is not entirely exhaustive and does not claim to include all available interviews.
Spanning the years 2020 to 2024, the dataset offers a detailed perspective on political discourse during two legislative terms of the German Bundestag: the 19th term (2017–2021) and the 20th term (2021–2024).

\subsection{Transcription and Validation} The audio content of the interviews was processed through a systematic pipeline that included both computational methods and human validation. Audio files were transcribed using the Whisper transcription model, which supports German-language audio and produces high-quality segment-based transcriptions with timestamps \citep{radford2022whisper}.
We performed speaker diarization using embeddings extracted from each audio segment to identify and differentiate speakers within the interviews. A pre-trained ECAPA-TDNN model \citep{DBLP:conf/interspeech/DesplanquesTD20} processed each segment, and agglomerative clustering was applied to assign speaker labels.

To ensure the quality of the transcriptions, all transcripts were manually reviewed and corrected by two research assistants. The final transcripts were saved in a tabular format, with timestamps and speaker IDs clearly defined for each sentence. This rigorous combination of automated and manual processes ensured the dataset met the quality standards necessary for advanced natural language processing tasks.

\section{Data Overview} 

\subsection{Variables}

Below, we outline each of the variables in the main dataset.  

\begin{itemize}

\item The variable \textbf{format\_id} stores the unique identifier for each interview format (e.g., ARDSO for ARD Sommerinterviews). 

\item The variable \textbf{year} represents the year in which the interview was conducted. 

\item The variable \textbf{video\_id} stores the unique identifier for each video. This is particularly useful for linking back to the original video source or performing analysis at the video level.

\item The variable \textbf{speaker} contains a unique identifier for each speaker (e.g., a politician or interviewer). Interviewers are coded as Q999, while politicians are identified using their Wikidata ID (e.g., \href{https://www.wikidata.org/wiki/Q61053}{Q61053} for Olaf Scholz). This identifier can easily be used to map the speaker to other relevant data sources, such as LegislatoR \citep{gobel2022comparative} or Parlspeech \citep{rauh2020parlspeech} database.

\item The variable \textbf{text} contains the transcription of each interview segment as a character vector. This field forms the core content of the dataset, capturing the spoken words of politicians and interviewers. Researchers can analyze these transcriptions for various natural language processing (NLP) tasks, such as sentiment analysis, topic modelling, or linguistic style comparisons.

\item The variable \textbf{timestamp} captures the precise timing of each transcribed segment in the HH:MM:SS format. This temporal metadata facilitates the reconstruction of the speech sequence within an interview, allowing for dynamic analyses of conversational patterns, such as interruptions or shifts in dialogue. However, during the manual validation process, some timestamps were removed and are therefore only partially available.

\item The variable \textbf{order\_id} captures the position of each text within each video.

\end{itemize}

\subsection{Descriptive statistics}

The resulting dataset encompasses a variety of interview formats. Table \ref{table:formats_summary} presents an overview, including the format names, detailed descriptions, and links to their official webpages.

\renewcommand{\arraystretch}{1.5} 
\begin{table}[!htbp]
\centering
\caption{Formats}
\label{table:formats_summary}
\resizebox{\textwidth}{!}{%
\begin{tabular}{@{\extracolsep{5pt}}l|l|l|p{10cm}|l}
\hline 
\textbf{Format ID} & \textbf{No. Videos} & \textbf{Official Name} & \textbf{Description} & \textbf{Format Link} \\
\hline
MAISC & 35 & Maischberger & Interviews hosted by journalist Maischberger, tackling relevant current topics. & \href{https://www.daserste.de/information/talk/maischberger/index.html}{Link} \\
\hline
FRAGS & 28 & Frag Selbst & Interview based on YouTube comment questions & \href{https://www.tagesschau.de/thema/frag_selbst}{Link} \\
\hline
ARDSO & 20 & ARD Sommerinterviews & Televised annual interviews discussing pressing current affairs with key figures. & \href{https://www.ardmediathek.de/tagesschau24/sammlung/ard-sommerinterviews-und-frag-selbst-2024/ba9f6ead-0380-4bb6-bc70-8d61bd3639c3}{Link} \\
\hline
CARMI & 6 & Caren Miosga & Interviews hosted by journalist Caren Miosga, tackling relevant current topics. & \href{https://www.daserste.de/information/talk/caren-miosga/index.html}{Link} \\
\hline
ZDFWE & 10 & ZDF (Other videos) & Various rather short interviews with key politicians & \href{https://www.zdf.de/}{Link} \\
\hline
\end{tabular}%
}
\end{table}

\begin{figure}
    \centering
    \includegraphics[width=0.9\linewidth]{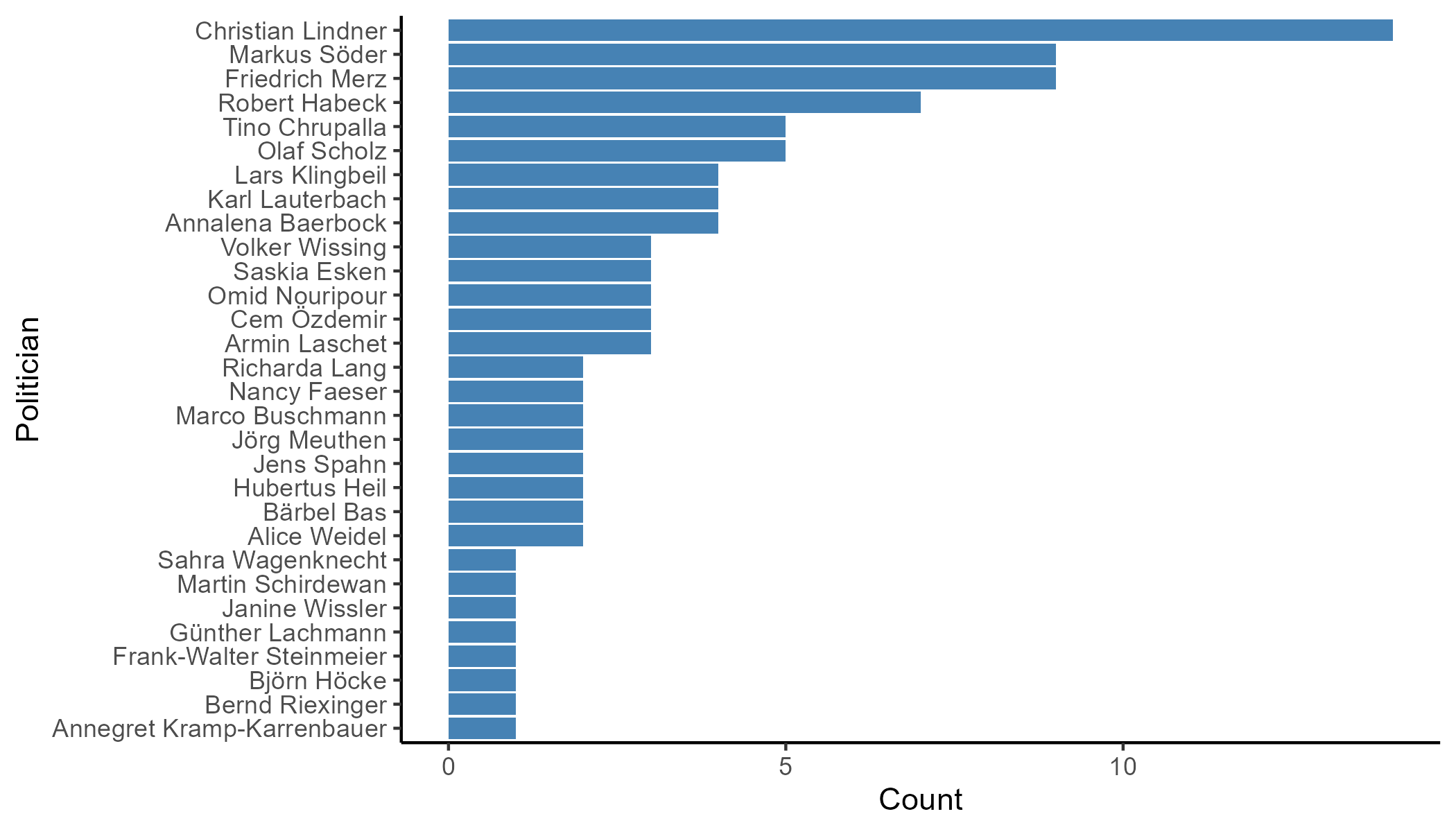}
    \caption{Summary Politicians}
    \label{pol}
\end{figure}

Figure \ref{pol} illustrates the distribution of the 33 politicians included in the dataset, with Christian Lindner (FDP) appearing most frequently, followed closely by Markus Söder (CSU), Friedrich Merz (CDU), and Robert Haback (Greens). 

\begin{figure}[!h]
    \centering
    \includegraphics[width=0.9\linewidth]{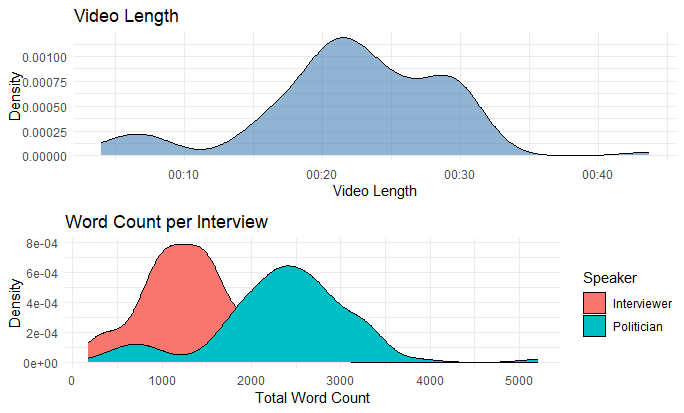}
    \caption{Data Summary}
    \label{speak}
\end{figure}

Figure \ref{speak} presents two density plots. The upper plot shows the distribution of video lengths, where most videos are around 20 to 30 minutes long, with a smaller number of videos reaching around 40 minutes. The lower plot illustrates the distribution of total word counts per interview. Politicians generally have a higher total word count (mean: 2360 words) compared to interviewers (mean: 1257 words), indicating that politicians speak more during interviews.

\section{Outlook}

The presented dataset lays the foundation for a broad range of future research opportunities in the domain of political communication and media studies. By offering systematically collected and transcribed interviews, it facilitates quantitative and qualitative investigations into agenda-setting, linguistic strategies, self-presentation, and the dynamics of interviewer-interviewee interactions.

Future expansions or enrichment of this dataset could include interviews beyond 2024 to enable longitudinal studies and capture evolving trends in political discourse. Additionally, integrating external metadata, such as audience reach or media outlet biases, could provide further context for analyzing the impact of these interviews. Advances in computational techniques, such as emotion recognition and semantic similarity measures, also open new avenues for examining deeper conversational patterns and sentiment dynamics.

Moreover, adapting this approach to other national contexts or extending it to cover non-German-speaking political environments could yield valuable comparative insights. The dataset thus serves as a stepping stone for exploring political communication within and beyond Germany, contributing to a more comprehensive understanding of the mediatization of politics in the digital era.

\bibliographystyle{johd}
\bibliography{bib}

\end{document}